\documentclass[11pt,a4paper]{article}
\usepackage[hyperref]{emnlp2020}
\usepackage{times}
\usepackage{latexsym}

\usepackage{xspace}
\usepackage{graphics}
\usepackage{graphicx}
\usepackage{booktabs}
\usepackage{enumitem}
\usepackage{comment}
\usepackage{array}
\usepackage{amsmath}
\usepackage{amssymb}

\usepackage{microtype}

\aclfinalcopy 

\newcommand\numsubfigs{2069\xspace}
\newcommand\numsubcaps{7507\xspace}

\newcommand\semscholar{\texttt{anonymized}\xspace}

\newcommand{\medicate}{\textsc{MedICaT}\xspace}

\newcommand{\expnumber}[2]{{#1}\mathrm{e}{#2}}

\title{MedICaT: A Dataset of Medical Images, Captions, and Textual References}

\author{Sanjay Subramanian$^1$ \quad
Lucy Lu Wang$^1$ \quad
Sachin Mehta$^2$ \quad
Ben Bogin$^3$ \quad
Madeleine van \\ 
\textbf{Zuylen}$^1$ \quad
\textbf{Sravanthi Parasa}$^4$ \quad
\textbf{Sameer Singh}$^5$ \quad
\textbf{Matt Gardner}$^1$ \quad
\textbf{Hannaneh Hajishirzi}$^{1,2}$
\\ [2mm]
  $^1$
  Allen Institute for AI \quad
  $^2$
  University of Washington \quad
  $^3$
  Tel Aviv University \\
  $^4$
  Swedish Medical Group \quad
  $^5$
  University of California, Irvine \\
  {\tt \{sanjays, lucyw\}@allenai.org}
}

\date{}

\begin{document}
\maketitle
\begin{abstract}
Understanding the relationship between figures and text is key to scientific document understanding. 
Medical figures in particular are quite complex, often consisting of several subfigures (75\% of figures in our dataset), with detailed text describing their content. 
Previous work studying figures in scientific papers focused on classifying figure content rather than understanding how images relate to the text.
To address challenges in figure retrieval and figure-to-text alignment, we introduce \medicate, a dataset of medical images in context. \medicate consists of 217K images from 131K open access biomedical papers, and includes captions, inline references for 74\% of figures, and manually annotated subfigures and subcaptions for a subset of figures.
Using \medicate, we introduce the task of subfigure to subcaption alignment in compound figures and demonstrate the utility of inline references in image-text matching.
Our data and code can be accessed at  \href{https://github.com/allenai/medicat}{https://github.com/allenai/medicat}.
\end{abstract}

\section{Introduction}
\label{sec:intro}

Scientific document understanding necessitates the analysis of various components of scientific papers, including recognizing relationships between the text, figures, and references of papers. In the medical domain, connections between the text and figures in a paper are useful to enable the retrieval of figures via textual queries and to produce systems that are capable of analyzing and understanding medical images.

\begin{figure}[tb!]
\centering
  \includegraphics[width=\linewidth]{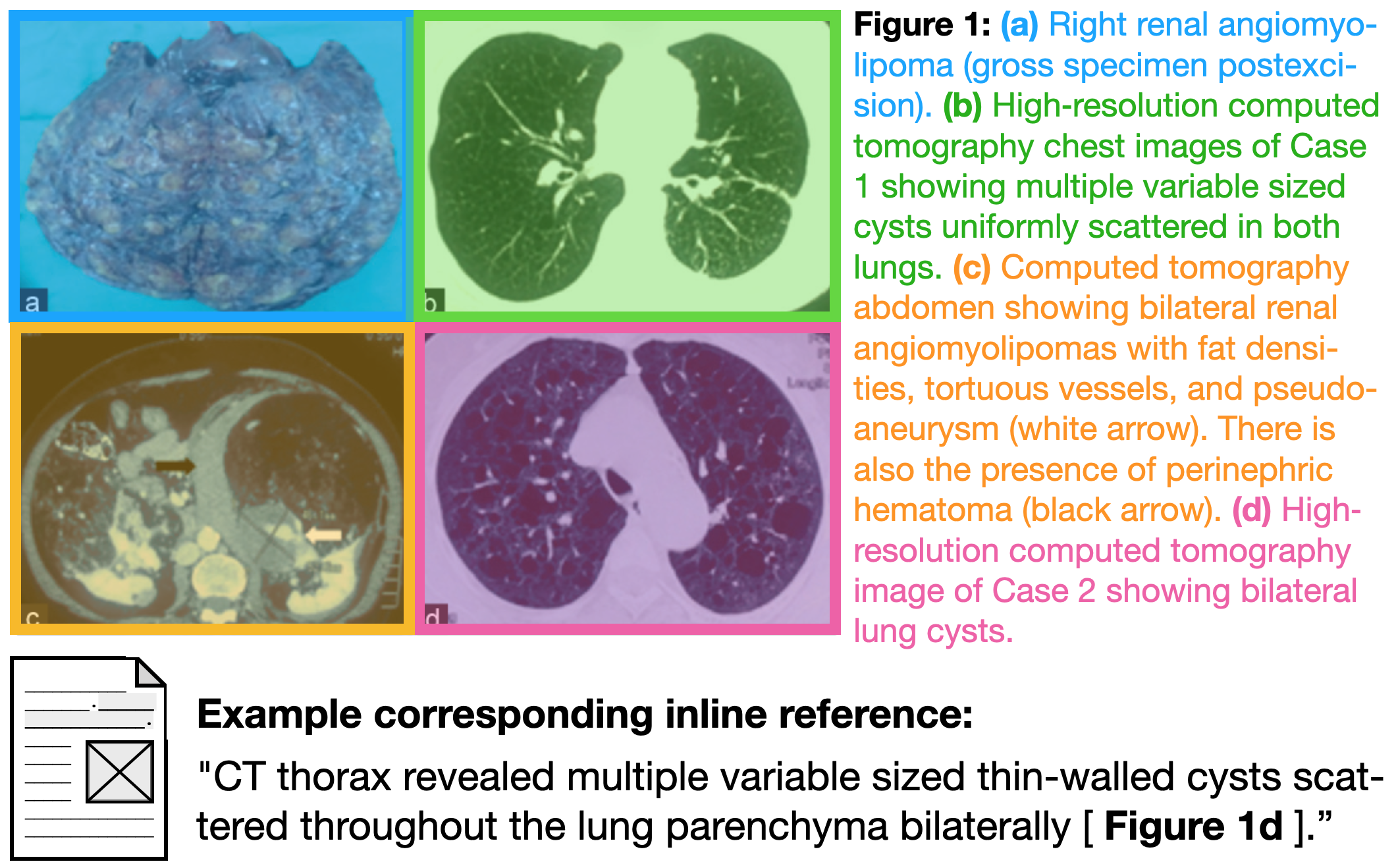}
  \caption{Example of color-coded subfigures and corresponding subcaptions (\emph{top}), with an example inline reference from the full text. Figure and text adapted from \citet{Dhungana2018RareCL}.\protect\footnotemark}
  \label{fig:pair-caption}
\end{figure}

Modern academic search engines are able to extract figures and associated captions from papers, so that queries can be matched against captions to retrieve relevant images. However, scientific figures often contain subfigures (40\% of figures in PubMed \citep{de2016overview}), and for a given query, only some of these may be relevant. Consider a user searching ``lung cyst CT''; ideally, the user would be shown just parts (b) and (d) from Figure~\ref{fig:pair-caption}.
But selecting relevant subfigures requires finding and aligning subcaptions with subfigures, and current systems lack this ability. Existing large-scale datasets \citep{Pelka2018RadiologyOI,ImageCLEF18} explicitly exclude compound figures, while datasets with subfigure annotations \citep{de2016overview,you2011detecting} do not provide annotations for aligning subfigures with subcaptions, and therefore cannot support subfigure retrieval.

\footnotetext{\href{https://creativecommons.org/licenses/by-nc-sa/4.0/}{https://creativecommons.org/licenses/by-nc-sa/4.0/}}

Images and textual descriptions in medical papers are also useful for developing systems for automated medical image analysis. Previous work \citep{ImageCLEF18,ImageCLEFmedConceptOverview2019} has used such data for tasks such as captioning and concept tagging, but the textual descriptions used in their work are limited to captions. Important details about figures are also often available in the main body of a paper, as in Figure~\ref{fig:pair-caption}, where an inline reference provides additional context that the CT is of the thorax and that the cysts seen are thin-walled. These inline references, if available, are useful for training medical image analysis systems. 

We introduce \medicate, a dataset of medical figures, captions, subfigures/subcaptions, and inline references that enables the study of these figures in context.
Our dataset includes compound figures (75\% of figures) with manual annotations of subfigures and subcaptions (\numsubcaps subcaptions for \numsubfigs compound figures). Using \medicate, we introduce the task of subfigure-subcaption alignment, a challenging task due to high variation in how subfigures are referenced.
We provide a strong baseline model based on a pre-trained transformer that obtains an F1 score of 0.674, relative to an estimated inter-annotator agreement of 0.89. \medicate is also richer than previous datasets in that it includes inline references for 74\% of figures in the dataset. We show that training on references in addition to captions improves performance on the task of matching images with their associated descriptions. 
By providing this rich set of relationships between figures and text in papers, our dataset enables the study of scientific figures in context.
\section{The \medicate Dataset}
\label{sec:dataset}

We extract figures and captions from open access papers in PubMed Central
using the results of \citet{Siegel2018ExtractingSF}. To add inline references, we match extracted figures to corresponding figures in the publicly-available S2ORC corpus \citep{Lo2020S2ORCTS}, then extract inline references for these figures from the S2ORC full text (see Appendix~\ref{app:inline} for details). We exclude figures found in ROCO \citep{Pelka2018RadiologyOI} to create a disjoint dataset, though we identify and release inline references for ROCO figures.

\paragraph{Medical image filters}
We define medical images as those that visualize the interior of the body for clinical or research purposes. This includes images generated by radiology, histology, and other visual scoping procedures. To identify medical images, we first use a set of keywords describing medical imaging techniques (such as \emph{MRI} or \emph{ultrasound} -- see Appendix~\ref{app:keywords}) to match across the caption and reference text, discarding images where a keyword does not appear. 
After filtering by keyword, some images are still non-medical, e.g., some are natural images of medical imaging equipment, or graphs showing image-derived measurements. To remove non-medical images, we apply an image classifier (ResNet-101 \citep{he2016deep} pretrained on ImageNet \citep{Deng2009ImageNetAL}) similar to \citet{ImageCLEF18}. The classifier is trained on the DocFigure dataset \citep{jobin2019docfigure}, which contains 33K figures from scientific articles annotated to 28 classes like ``Medical'' or ``Natural'' images . We keep an image if ``Medical'' is in the top $K=4$ labels predicted by the classifier (precision: 96\%, recall: 67\%).\footnote{$K$ is tuned via manually annotating a set of 200 randomly selected images from the keyword-filtered results, which is also used to compute the provided precision and recall.}

\begin{table}[tb!]
  \small
  \centering
  \begin{tabular}{lr}
    \toprule
       \bf Dataset Statistics & \bf \medicate \\
    \midrule
       Number of papers & 131,410 \\
       Number of figures & 217,060 \\
       Avg. figures per paper & 1.7 \\ 
       Avg. inline refs. per figure & 1.4 \\ 
       Avg. caption length (tokens) & 74.2 \\ 
       Pct. with reference text & 74\% \\ 
       Avg. reference length (tokens) & 67.3 \\ 
       Avg. Jaccard similarity btwn cap. and ref. & 23\% \\ 
       Pct. medical figures* & 93\% \\
       Pct. with subfigures* & 75\% \\
    \bottomrule
  \end{tabular}
  \caption{Dataset statistics. Items marked with * are estimated on a sample of 2327 figures by four annotators.}
  \label{tab:s2_stats}
\end{table}

\paragraph{Dataset statistics}
Table~\ref{tab:s2_stats} provides statistics on \medicate. We note that the images in \medicate are diverse; through manual assessment of 200 images, we estimate that the dataset contains primarily radiology images (72\%), along with histology images (13\%), scope procedures (3\%), and other types of medical images (7\%).

\paragraph{Manual Annotations}
We collect bounding boxes for subfigures and annotations of corresponding subcaptions for \numsubfigs figures, resulting in \numsubcaps subfigure-subcaption pairs. These annotations were collected in two phases. In the first phase, five annotators (with various degrees of biomedical training, ranging from none to graduate level, though no annotators are medical doctors) marked subfigures in each figure and wrote single-span subcaptions for each subfigure when possible. To compute inter-annotator agreement, three annotators annotated the same set of 100 figures. An agreement score is calculated by taking every ordered pair among these three annotators, treating one as gold and one as predictor, and computing the metric described in \S\ref{sec:subcaption-task}. An average is computed over all pairs, giving an agreement score of 0.828. In the second phase, two of these annotators reviewed the existing annotations and revised subcaptions (and in some cases, subfigures) with the option of having multi-span subcaptions. The inter-annotator agreement in the second phase, computed on 100 figures, is 0.89.\footnote{For the annotator agreement calculation in the second phase, annotators were provided the subfigures annotated in the first phase but were not provided any subcaptions written in the first phase.}
\section{Subfigure and Subcaption Alignment}
\label{sec:subcaption-task}

A major challenge of scientific figure understanding is the prevalence of compound figures (around 75\% in \medicate are compound). As discussed in Section \ref{sec:intro}, matching subfigures with their corresponding subcaptions can be useful for image retrieval. A potential use case is seen in \citet{wang2020covid}, who use a subfigure-subcaption alignment model to extract relationships for a COVID-19 knowledge graph.
Though compound figure segmentation is studied by \citet{de2016overview}, they only assign to each subfigure a label describing image modality/type, ignoring other information in subcaptions. \citet{you2011detecting} build a dataset and system to detect subfigure labels (e.g. A/B/C) but ignore other ways in which subfigures are referenced in captions (e.g. spatial position), and their dataset is small (515 figures).

\begin{figure}[tb!]
    \centering
    \includegraphics[width=\linewidth]{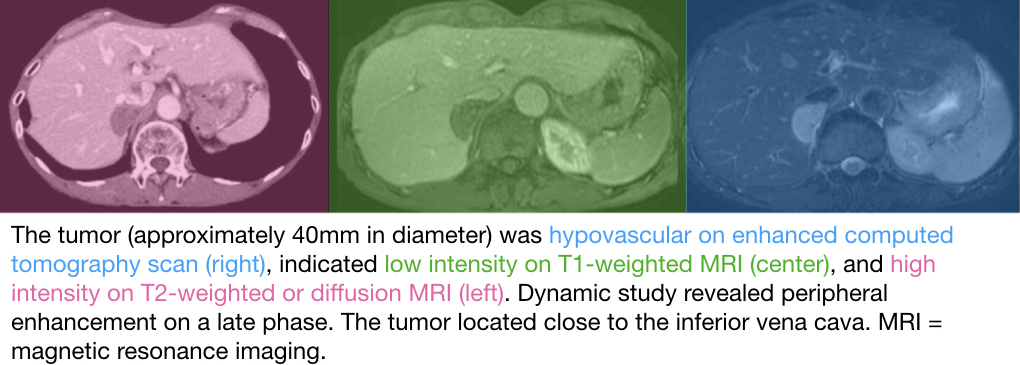}
    \caption{Subfigure to subcaption alignment is challenging for this figure because the subfigures are referenced by spatial position in the right-to-left order. Corresponding subfigures and subcaptions are indicated by color. Figure and caption adapted from \citet{Ohkura2015PrimaryAL}.\footnote{https://creativecommons.org/licenses/by-nc-nd/2.0/}}
    \label{fig:challenging_example}
\end{figure}

\paragraph{Task} Motivated by this problem, we propose the task of subfigure to subcaption alignment. Given a possibly compound figure and its caption, the task entails identifying (1) each subfigure and (2) the corresponding subcaption for each subfigure. We define a subfigure's subcaption to be the set of tokens in the caption that reference/describe the subfigure but do not describe all subfigures.\footnote{We also typically exclude scale/measurement information.} This task is challenging because subfigures are referenced in a variety of ways in our dataset, e.g., described in pairs or groups, referenced by spatial position (e.g. \emph{upper left} or \emph{second column}), or in multiple subcaption spans within the same caption. Figure~\ref{fig:challenging_example} displays a challenging example.

\paragraph{Model} For subfigure detection, we use the Faster R-CNN object detector \citep{ren2015faster} with a ResNet-50 \citep{he2016deep} backbone pre-trained on ImageNet \citep{Deng2009ImageNetAL}. For figures without subfigure annotations, we set the gold annotation as a single bounding box over the entire figure.

We propose and implement two models for subcaption extraction and alignment. In both models, a CRF is applied to the output of a BERT encoder \citep{huang2015bidirectional,devlin2019bert}. This model is used frequently for named entity recognition (NER), and we use it here to extract spans from the caption.\footnote{In NER, the model must also predict the type of each span; here we treat each span as having the same type.} The first model (Text-only CRF) segments the caption into subcaptions with only the caption as input, then heuristically aligns the subcaptions to detected subfigures. To train this model, we iterate over gold subcaptions and extract for each subcaption the longest sub-span that does not overlap with spans extracted for previous subcaptions. After extracting subcaptions, we heuristically match subcaptions with subfigures as follows. We sort subcaptions by the order in which they occur in the caption. We sort subfigures first by vertical position (row), then by horizontal position (column).\footnote{Two subfigures are in the same row if the vertical coordinates of their top left corners differ by $<$ 50 pixels.} We pair each subfigure with the corresponding subcaption in this sorted order. If the number of predicted subfigures exceeds the number of predicted subcaptions, we use the last predicted subcaption for all remaining subfigures.

In the second approach (Text+Box Embedding CRF), the model takes as input a subfigure bounding box, which is projected to a box embedding and concatenated with the token encodings produced by BERT. The tag probabilities for each token are predicted by a multi-layer perceptron over these concatenated encodings. Since this model predicts subcaptions separately for each subfigure, it is capable of extracting multi-span subcaptions, in contrast to the first model.

\paragraph{Evaluation} We find for each gold subfigure $ G $ the predicted subfigure $ P $ that maximizes $ IOU(G, P) $, where $ IOU(\cdot, \cdot) $ denotes the intersection-over-union between two regions \citep{everingham2010pascal}. If the IOU is less than the threshold $ 0.5 $, the model obtains a score of $ 0 $ for $ G $. If the IOU exceeds $ 0.5 $, the model's score for $ G $ is equal to the F1 between the set of tokens in the gold subcaption for $ G $ and the set of tokens in the predicted subcaption for $ P $ ignoring non-alphanumeric tokens. Gold subfigures without subcaptions are excluded from evaluation. The overall score is defined as the average over the scores of all gold subfigures. We also report mean average precision (mAP) for subfigure detection based on the COCO \citep{lin2014microsoft} evaluation.

\begin{table}[tb!]
  \small
  \centering
  \begin{tabular}{p{47mm}p{11mm}p{6mm}}
    \toprule
      \bf Model & \bf Validation F1 & \bf Test F1 
 \\
    \midrule
    Oracles + alignment heuristic & 88.0 & 90.8 \\
    Oracles (single-span) + alignment heuristic & 78.1 & 75.5 \\
    \midrule
    \textbf{Gold subfigure oracle} & & \\ [0.5mm]
      Text-only CRF $_\text{\textrm{w/o pretrained weights}}$ & 47.8 & 43.3 \\
      Text-only CRF $_\text{\textrm{w/ pretrained weights}}$ & 71.0 & 66.9 \\
      Text+Box Embedding CRF & \textbf{74.1} & \textbf{71.9} \\
     \midrule
    \textbf{Using predicted subfigures} & & \\ [0.5mm]
      Text-only CRF $_\text{\textrm{w/o pretrained weights}}$ & 44.7 & 40.3 \\
      Text-only CRF $_\text{\textrm{w/ pretrained weights}}$ & 66.4 & 61.3 \\
      Text+Box Embedding CRF & \textbf{69.9} & \textbf{67.5} \\
    \bottomrule
  \end{tabular}
  \caption{Results for subfigure-subcaption alignment. \emph{Oracles + alignment heuristic} uses gold subfigures and gold subcaptions with the alignment heuristic. \emph{Oracles (single-span) + alignment heuristic} uses gold non-overlapping single-span subcaptions extracted as in the training of the SciBERT CRF model.}
  \label{tab:subcaption_results}
\end{table}

\paragraph{Experiment results} We split our data into train (65\%), validation (15\%), and test (20\%) sets randomly.\footnote{All examples used for computing the annotator agreement score in the second annotation phase are placed in the test set.}
For subfigure detection, we obtain a mAP score of 79.3 on the test set.
For subcaption extraction, we use SciBERT \citep{Beltagy2019SciBERTPC} tokenization because compared to the vanilla BERT vocabulary, the SciBERT vocabulary includes more of the words in the captions, resulting in a smaller number of wordpieces when using SciBERT tokenization. We also experiment with initializing the BERT encoder with SciBERT pre-trained weights. See Appendix \ref{app:subcaption} for further details such as hyperparameter tuning.

Table~\ref{tab:subcaption_results} shows results for our baseline models on the subfigure-subcaption alignment task. We report results with a gold subfigure oracle to separate the error caused by subfigure detection from that caused by subcaption extraction. Initialization with SciBERT pre-trained weights improves performance considerably, consistent with previous results on various biomedical NLP tasks. The maximum achievable performance with the alignment heuristic (using gold subcaptions and gold subfigures) is also given, indicating that alignment accounts for a large portion of the error. The oracle performance with single-span subcaptions (of the kind that can be predicted by the CRF model with the heuristic) is far below the unconstrained oracle performance, showing that a substantial number of subfigures require multi-span subcaptions. Finally, the box embedding model consistently outperforms the models using single-spans and the alignment heuristic.

\paragraph{Error Analysis} To understand the sources of error in the Text+Box Embedding CRF with subfigure oracle, we analyze 50 subfigures in the validation set for which the system obtains F1 $<0.5$. Most errors fall into these categories (not mutually exclusive): (a) predicted subcaption describes a different subfigure (46\%), (b) predicted subcaption is empty (22\%), (c) missing words in the predicted subcaption (14\%), and (d) annotation errors (6\%). Type (a) errors indicate that alignment (as opposed to subfigure/subcaption segmentation) is a major source of error.
\section{Image-text matching}
\label{sec:retrieval}

To demonstrate the utility of inline references in \medicate, we conduct experiments in image-text matching \citep{hodosh2013framing}, a task that has been studied extensively in the domain of natural images. Given a piece of text, the system's goal is to return the matching image from a database. Like image captioning, this task assesses the model's ability to align the image and text modalities, but the matching task avoids the issue of evaluating generated text \citep{hodosh2013framing}.

In the context of medical figures and captions from papers, we attempt to retrieve a corresponding figure given its caption. Since our dataset provides more than one textual description of each image via inline references, we analyze the benefit of using references as additional training data, and demonstrate improvements over models trained on captions only. At test time, only captions are used to enable a fair comparison of the models. We use a model similar to \citet{chen2019uniter} with a different token type embedding for inline references.

\begin{table}[tb!]
  \small
  \centering
  \begin{tabular}{p{20mm}llll}
 \toprule
   \bf Model & \bf R@1 & \bf R@5 & \bf R@10 & \bf R@20 
 \\
 \midrule
   Captions & $7.6_{0.59}$ & $26_{1.6}$ & $41_{1.6}$ & $58_{1.3}$ \\ [1mm]
   Caps + Refs & $9.4_{0.38}$ & $30_{1.2}$ & $44_{1.7}$ & $60_{1.2}$  \\ 
 \bottomrule
  \end{tabular}
  \caption{Results for image-text matching. Results show \emph{mean} percent accuracy with \emph{error} in subscript over $n=5$ random seeds. The \emph{error} is the standard error $ \sigma/\sqrt{n} $, where $ \sigma $ is the standard deviation over random seeds.}
  \label{tab:retrieval_results}
\end{table}

\paragraph{Linking with ROCO}
ROCO is a dataset of non-compound radiology figures and captions extracted from literature \citep{Pelka2018RadiologyOI}.\footnote{The ROCO dataset can be accessed at \href{https://github.com/razorx89/roco-dataset}{https://github.com/razorx89/roco-dataset}.} The ROCO dataset consists of around 82K figures (train/validation/test splits: 65K, 8175, 8177 respectively). We identify papers associated with figures in the ROCO dataset and extract the associated inline references, and release these as part of \medicate. We find inline references for approximately 25K figures in the ROCO dataset (approximately 21K in train and 4K each in validation and test splits). 

\paragraph{Model} Our model follows \citet{chen2019uniter} with a few modifications. We tokenize the input text and pass the token embeddings through a BERT encoder \citep{devlin2019bert}. The BERT encoder is initialized with SciBERT weights, and we use SciBERT uncased tokenization \citep{Beltagy2019SciBERTPC}. Pre-trained weights improve performance on the image-text matching task considerably. We insert the visual representations, projected into the hidden state dimensionality, as extra hidden states in the middle of the encoder (layer 6). In contrast to \citet{chen2019uniter}, we do not use an object detector to find regions of interest in the image, since objects tend to be sparse in medical images. Instead, the visual representation is obtained by an affine transformation of the feature vector produced for the entire image by a ResNet-50 network pretrained on ImageNet.\footnote{We also add another embedding to this image representation (similar to the position embedding for tokens).} The other noteworthy difference in our model is that we use different token type embeddings for inline references and captions, since inline references are different in style and content from captions. During training, for each piece of text, the model is given the correct image as well as two other images (negative images) sampled uniformly at random from all images. The training objective is choosing the correct one of these three images. The validation accuracy is measured on the same task, except 20 negative images are sampled for each piece of text. Details on hyperparameter tuning and training are provided in Appendix \ref{app:retrieval}.

\paragraph{Experiment results} Experiments are performed on  ROCO using the provided train/validation/test splits \citep{Pelka2018RadiologyOI} (Table~\ref{tab:retrieval_results}). We present results on 2000 randomly sampled test set image/caption pairs due to the time complexity of evaluation. As in previous work, we use the Recall@K metric, which gives the proportion of examples for which the system's rank of the correct image is in the top K. Training with references improves upon training with captions alone. Since we were only able to extract associated inline references for 33\% (21K) of the training images, we expect a greater improvement if references were available for more figures. As documented in Appendix \ref{app:retrieval}, the minimum and maximum performance over random seeds are also higher when training with both captions and references than when training with captions only.
\section{Discussion}
\label{sec:discussion}

\medicate allows medical figures to be studied in the context of their source papers by providing links between subfigures and subcaptions and between figures and inline references. We propose the task of subfigure-subcaption alignment and provide a strong baseline model, and we demonstrate the utility of inline references for image-text matching.

\medicate can also benefit other medical vision-language tasks (e.g. captioning, VQA). Pre-training techniques that have worked well for these problems in the general domain by using large numbers of aligned text and images \citep{zhou2019unified} can leverage the aligned data in \medicate. The techniques we use to construct \medicate can also be extended beyond ``medical images'' to study the relationships between figures and text in scientific documents from other domains.

\section*{Acknowledgments}
 We thank P.V. Subramanian for his help with annotations. We thank the Semantic Scholar and AllenNLP teams at AI2, UW-NLP, and H2lab at UW for helpful comments and feedback throughout the data collection and writing process. We would like to acknowledge grants from ONR MURI N00014-18-1-2670 and NSF (IIS 1616112, NSF IIS 1817183).

\bibliography{imgs}
\bibliographystyle{acl_natbib}

\appendix
\section{Inline reference extraction}
\label{app:inline}

\begin{figure}[b!]
  \centering
  \includegraphics[width=\linewidth]{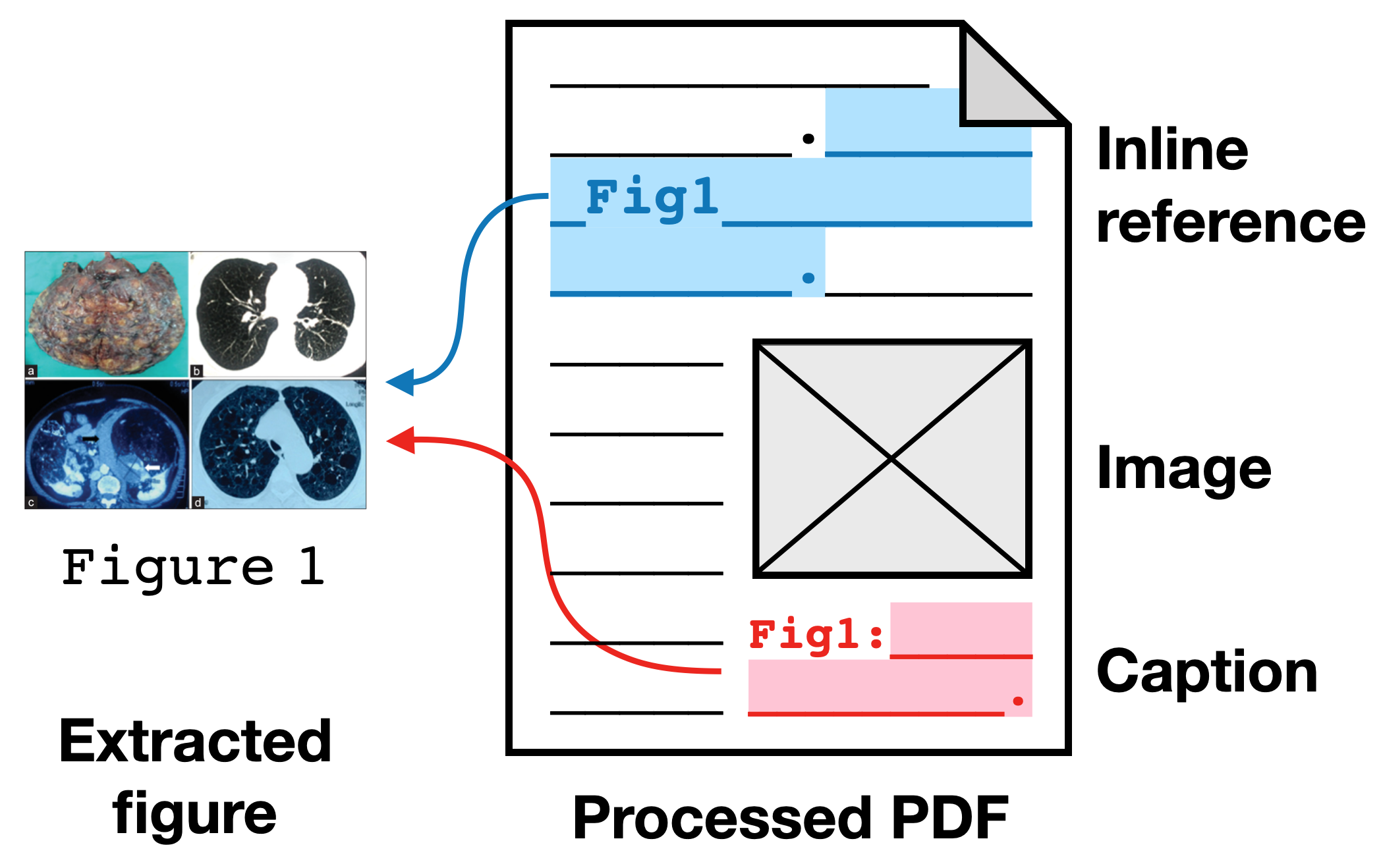}
  \caption{Extracted figures from \semscholar are aligned with the S2ORC parse of the paper PDF to link caption text (\emph{red}) and inline references (\emph{blue}) to each image. Example figure from \citet{Dhungana2018RareCL}}
  \label{fig:schematic}
\end{figure}

We define each inline reference as the sentence from the full text of the paper that makes reference to a figure object (see Figure~\ref{fig:schematic}). We extract inline references from the S2ORC dataset, a publicly available dataset of 8M+ full text papers \citep{Lo2020S2ORCTS}. References to figure or table objects in the full text of S2ORC are annotated, and we leverage this feature to extract inline references and link them to figures.

We begin with the set of figures and captions extracted from open access papers in \semscholar. We then identify corresponding papers in S2ORC using paper identifiers such as DOI or PMC ID. We extract all figure captions and inline references from S2ORC for these corresponding papers, using scispaCy \citep{Neumann2019ScispaCyFA} to identify sentence boundaries for inline references. 

Both the \semscholar and S2ORC corpuses have figure captions, while only \semscholar contains images and only the S2ORC corpus contains inline references. To identify inline references, \semscholar and S2ORC data must be matched based on figure caption. Captions are matched based on extracted figure index (e.g. \emph{Figure 1} or \emph{Fig. 2}) and token Jaccard overlap between caption text. When the figure index is available in both caption extractions and are the same, this designates a match. When figure index is not available, captions are matched if the token Jaccard between them is greater than 0.8. Once the two datasets are aligned in this fashion, we append the S2ORC reference for each figure to the corresponding figure extraction from \semscholar to create \medicate.

\section{Medical image filter keywords}
\label{app:keywords}

A set of keyword filters are used as a first pass for identifying medical images. Because of the large size of the initial \semscholar figure extraction dataset, which contains many millions of images, it is impractical to run the medical image classifier on all extracted figures. Keyword filters act to select medical images with lower precision but adequate recall, to then be input to the medical image classifier. 

In conference with a medical doctor, common terms describing medical images are identified as keywords. Figures whose captions and references match against a keyword (case-insensitive) are kept. The full set of keywords used is provided below:

\begin{itemize}[noitemsep]
  \item[] MRI \quad fMRI \quad CT
  \item[] CAT \quad PET \quad PET-MRI
  \item[] MEG \quad EEG \quad ultrasound
  \item[] X-ray \quad Xray \quad nuclear
  \item[] imaging \quad tracer \quad isotope
  \item[] scan \quad positron \quad EKG
  \item[] spectroscopy \quad radiograph
  \item[] tomography \quad endoscope
  \item[] endoscopy \quad colonoscopy
  \item[] elastography \quad ultrasonic
  \item[] ultrasonography \quad echocardiogram
  \item[] endomicroscopy \quad pancreatoscopy
  \item[] cholangioscopy \quad enteroscopy
  \item[] retroscopy \quad chromoendoscopy
  \item[] sigmoidoscopy \quad cholangiography
  \item[] pancreatography
  \item[] cholangio-pancreatography
  \item[] esophagogastroduodenoscopy
\end{itemize}
\section{Subfigure-subcaption alignment model}
\label{app:subcaption}

In this section, we give further experimental details for the subfigure-subcaption alignment task. When using predicted subfigure detections to align with the predicted subcaptions, we choose a confidence threshold of 0.7 for the Faster-RCNN predictions.

For the SciBERT+Box Embedding model, at test time, we compute a probability for each span by adding the model's computed probabilities of the start and end tokens of the span. Then we select the span with the highest probability among all \emph{valid} spans, where a valid span has an end token that does not precede the start token.

For all models we train with early stopping, where training is stopped when validation performance has not improved in the last five epochs. For subfigure detection models, we use the Adam optimizer \citep{kingma2014adam}, and we use a batch size of 10. We tune the learning rate and no other hyperparameters. The method used for hyperparameter search is random search. The learning rate is sampled from a log-uniform distribution over $ (\expnumber{1}{-5}, \expnumber{1}{-3}) $. We perform 10 trials for hyperparameter search and choose the model with the highest mAP score on the validation data, which has a learning rate of $ \approx \expnumber{2.22}{-4} $. The number of parameters in this model is 41.4M parameters.

For subcaption extraction models, we use the BERT Adam optimizer \citep{devlin2019bert}, and we use a batch size of 8. We tune the following hyperparameters: learning rate, weight decay (not applied to bias parameters or LayerNorm parameters), and dropout rate. Our method for hyperparameter search is random search, where learning rate is sampled from a log-uniform distribution over $ (\expnumber{5}{-6}, \expnumber{1}{-4}) $, weight decay is sampled uniformly from $ (0, 1) $, and dropout is sampled uniformly from $ (0, 0.5) $. For each model, we perform 30 trials for hyperparameter search. For the CRF Tagger models, the validation metric is the span F1 (precision is the proportion of predicted spans that occur in the gold subcaptions, and recall is the proportion of gold spans that occur in the predictions). For the box embedding model, the validation metric is the word F1 between the predicted subcaption for the given box and the gold subcaption for that box. These validation metrics were used to select hyperparameter choices and were used also for early stopping as described above. The hyperparameter settings that yielded the best performance for each model are given below. The number of parameters for each model is also provided.
\paragraph{CRF Tagger without SciBERT-pretrained Weights} Learning rate: $ \expnumber{2.47}{-5} $, Weight decay: $ 0.574 $, Dropout: $ 0.499 $, Number of parameters: 109.9M
\paragraph{CRF Tagger with SciBERT-pretrained Weights} Learning rate: $ \expnumber{2.60}{-5} $, Weight decay: $ 0.770 $, Dropout: $ 0.182 $, Number of parameters: 109.9M
\paragraph{SciBERT with Box Embedding} Learning rate: $ \expnumber{1.34}{-5} $, Weight decay: $ 0.699 $, Dropout: $ 0.404 $, Number of parameters: 221.1M

\paragraph{Computing infrastructure} Experiments are performed (1) on systems running Google Kubernetes Engine (container OS) that each have 16 CPUs, 104 GB of main memory, and 1 P100 GPU (16 GB memory), and (2) on a system running Ubuntu 18.04 that has 64 CPUs, 512 GB of main memory, and 8 RTX 8000 GPUs (48 GB memory). Only 1 GPU was used in each experiment.

\paragraph{Running Time} The following running times were obtained on the second type of system described above, each using a single RTX 8000 GPU. For each subcaption extraction, we give the time for predicting subcaptions on our validation set of 312 figures with a batch size of 1. These estimates include the time for loading data. For SciBERT with Box Embedding, recall that the model is run separately for each subfigure.

\noindent
CRF Tagger without SciBERT-pretrained Weights: 13 seconds

\noindent
CRF Tagger with SciBERT-pretrained Weights: 13 seconds

\noindent
CRF Tagger with Box Embedding: 51 seconds

The average prediction time for the subfigure detection model with a batch size of 1 is 77 seconds for our validation set of 316 figures.

For the final set of experiments, approximately 79.4 GPU hours were used for training the subfigure and subcaption models. Note that this amount includes hyperparameter tuning for the final set of experiments but does not include previous experiments that were done (e.g. during model development).
\section{Image-text Matching model}
\label{app:retrieval}

Here, we provide further experimental details for the image-text matching experiments. We use the Adam optimizer \citep{kingma2014adam}, with a batch size of 16.
We fix the learning rate to be $ \expnumber{1}{-5} $ and the dropout rate to be $ 0.1 $. Hyperparameter tuning was used to select the layer to insert the visual representation using data from the ImageCLEF-2019 VQA task \citep{ImageCLEFVQA-Med2019}. The tuning strategy is random search over 50 trials, where the layer number is sampled uniformly over the integers between $ 0 $ and $ 11 $ (inclusive).
Some manual tuning was done as well (about 30 trials). (In these tuning experiments, other hyperparameters (e.g. dropout) were varied as well, but these experiments did not determine the values of any hyperparameter other than the visual insert layer number.) The validation metric used to choose among the hyperparameter choices in these trials was accuracy on the VQA task.

Models are trained with early stopping, where training is stopped if the validation accuracy does not improve within five epochs. We use the same set of random seeds for both the model trained with captions only and the model trained with captions and references. Table~\ref{tab:retrieval_results_minmax} shows the results of the best and worst performing models of each of the two types (captions and captions \& references) over the five random seeds.

We use SciBERT \citep{Beltagy2019SciBERTPC} initialization in all of our models. We find that it yields better results on the image-text matching task in comparison to random initialization.

The model has 159.6M parameters
(for both the version trained on captions and that which is trained on captions+references, since the model architecture is the same). However, note that the model that is trained only on captions only makes use of one of the token type embeddings. (Each token type embedding has 768 parameters.)

\begin{table}[tb!]
  \small
  \centering
  \begin{tabular}{p{25mm}llll}
 \toprule
   Model & R@1 & R@5 & R@10 & R@20 
 \\
 \midrule
   Captions$_{max}$ & 9.4 & 31 & 46 & 61 \\
   Captions+Refs$_{max}$ & 10 & 32 & 47 & 63  \\
 \midrule
   Captions$_{min}$ & 5.8 & 21 & 35 & 53 \\
   Captions+Refs$_{min}$ & 8.3 & 25 & 38 & 57  \\
 \bottomrule
  \end{tabular}
  \caption{Results for image-text matching. The \emph{max} results provide the percent accuracy of the best model from the 5 training runs (each using different random seeds). Similarly, the \emph{min} results provide the percent accuracy of the worst model from the 5 training runs.}
  \label{tab:retrieval_results_minmax}
\end{table}

\paragraph{Computing Infrastructure} Experiments were run on a system running Ubuntu 18.04 that has 64 CPUs, 512 GB of main memory, and 8 RTX 8000 GPUs (48 GB memory). 1 GPU was used in each experiment.

\paragraph{Running Time} The average amount of time required to obtain predictions on the test set of 2000 instances is 122.1 minutes (including data loading time).

Training in the final set of 10 experiments for which we report results in this paper took approximately 343.7 GPU-hours. Note that this amount does not include other experiments done during the project (e.g. during model development).
\section{Annotation instructions}
\label{app:annotation}
Please see the PDF in \texttt{Supplementary Materials Data} for the instructions and examples that were provided to annotators for the first round of subfigure-subcaption annotations.

\end{document}